\documentclass[runningheads,a4paper]{llncs}
\usepackage{graphicx}
\usepackage{amsmath}
\usepackage{framed}
\usepackage{vntex}
\usepackage[english]{babel}
\usepackage{url}
\usepackage{here}

\begin{document}

\title{A Vietnamese Text-based Conversational Agent}

\author{Dai Quoc Nguyen\inst{1} \and Dat Quoc Nguyen\inst{1} \and Son Bao Pham\inst{1,2}}

\institute{Faculty of Information Technology \\
University of Engineering and Technology\\
Vietnam National University, Hanoi \\
\email{\{dainq, datnq, sonpb\}@vnu.edu.vn}
\and
Information Technology Institute \\ Vietnam National University, Hanoi 
}

\maketitle

\begin{abstract}
This paper introduces a Vietnamese text-based conversational agent architecture on specific knowledge domain which is integrated in a question answering system.
When the question answering system fails to provide answers to users' input, our conversational agent can step in to interact with users to provide answers to users.
Experimental results are promising where our Vietnamese text-based conversational agent achieves positive feedback in a study conducted in the university academic regulation domain. 

\end{abstract}

\section{Introduction}\label{Introduction}


A text-based conversational agent is a program allowing the conversational interactions between human and machine by using natural language through text.
The text-based conversational agent uses scripts organized into contexts comprising hierarchically constructed rules. 
The rules consist of patterns and associated responses, where the input is matched based on patterns and the corresponding responses are sent to user as output.

We focus on the analysis of input text in building  a conversational agent. 
Recently, the input analysis over user's statements have been developed following two main approaches: using keywords (ELIZA \cite{Weizenbaum:1983}, ALICE \cite{Richard:Alice}, ProBot \cite{Sammut:2001}), and using similarity measures (O'Shea et al. \cite{Oshea:2010}, Graesser et al. \cite{Graesser:2004}, Traum \cite{Traum:2006}) for pattern matching. 
The approaches using keywords usually utilize a scripting language to match the  input statements, while the other approaches measure the similarity between the statements and patterns from the agent's scripts. 

In this paper, we introduce a Vietnamese text-based conversational agent architecture on a specific knowledge domain.
Our system aims to direct the user's statement into an appropriate context.
The contexts are structured in a hierarchy of scripts consisting of rules in FrameScript language \cite{McGill:2003-8}.
In addition, our text-based conversational agent was constructed to integrate in a Vietnamese question answering system.
Our conversational agent provides not only information related to user's statement but also provides necessary knowledge to support our question answering system when it is unable to find an answer.

In section \ref{RelatedWorks}, we provide some related works about the text-based conversational agents, the FrameScript scripting language \cite{McGill:2003-8}, and present our overall conversational agent architecture in section \ref{Ourapproach}.  We describe our experiments and discussions in section \ref{Experiment}. The conclusion and future works will be presented in section \ref{Conclusion}.

\section{Related Works}\label{RelatedWorks}

\subsection{Text-based conversational agents}

ELIZA \cite{Weizenbaum:1983} is one of the earliest text-based conversational agents based on a simple pattern matching by identifying keywords from user's statement. 
ELIZA then transforms the user's statement into an appropriate rule and generates output response.

ALICE \cite{Richard:Alice} is a text-based conversational agent as chat robot utilizing an XML language called Artificial Intelligence Markup Language (AIML). 
AIML files consist of \textit{category} tags representing rules; each \textit{category} tag contains a pair of \textit{pattern} and \textit{template} tag. 
The system searches the \textit{pattern} according to user's input, and produces the appropriate \textit{template} as a response.
O'Shea et al. \cite{Oshea:2010} proposed a text-based conversational agent framework calculating the similarity between patterns from scripts and the user's input.
The highest ranked pattern is selected and its associated response is returned as output.
Graesser et al. \cite{Graesser:2004} presented a conversational agent called AUTOTUTOR matching input statements in the use of Latent Semantic Analysis. 
Traum \cite{Traum:2006} adapted the effective question answering characters \cite{Leuski:2006} to build a conversational agent also employing Latent Semantic Analysis for pattern matching.

Sammut \cite{Sammut:2001} presented a text-based conversational agent called ProBot that is able to extract data from users. 
ProBot's scripts are typically organized into hierarchical contexts consisting of a number of organized rules to handle unexpected inputs. 
Concurrently, McGill et al. \cite{McGill:2003-8} derived from ProBot's scripts \cite{Sammut:2001} build the rule system in FrameScript scripting language. 
FrameScript \cite{McGill:2003-8} provides the rapid prototyping of conversational interfaces and simplifies the writing of scripts.

\subsection{FrameScript Scripting Language}

FrameScript \cite{McGill:2003-8} is a language for creating a multi-modal user interfaces. It evolves from Sammut’s Probot \cite{Sammut:2001} to enable rule-based programming, frame representations and simple function evaluation. 

Each script in FrameScript includes a list of rules matched based on input statements and used to give an appropriate response. 
Rules are grouped into particular contexts of the form: \textit{context\_name :: rule\_set}.
The scripting rules in the FrameScript language consist of patterns and responses with the form: \textit{pattern ==$>$ response}. 

Every context is represented as a script, and a script is considered as a topic in a  domain. The domain is responsible for ensuring that the input statement is matched according to the correct scripts. A script has a \textit{trigger} to determine whether or not an input activates that context or topic. 
If the trigger does not exist, any input will activate the topic.
If an input is matched with a topic's trigger, the topic becomes the current context and the current topic.

Pattern expressions allow the use of the alternatives and existing pattern expressions.
Response expressions contain two different types: sequences and alternatives. 
Sequence of responses has a list of possible responses surrounded by brackets: 
[response 1 | response 2 $|$ ... $|$ another response]. 
Responses utilize the `$\#$' and `$^\wedge$' to perform actions such as to change the current context. 
For example, \textit{$\#$goto(a$\_$script)} or \textit{$\#$goto(a$\_$script, $<<$\textit{trigger}$>>$)} transforms a conversation or interaction from the current context to another one.
 If `$^\wedge$' is followed by an integer, the numbered pattern component associated with the integer is placed in the output response. 
 Some examples using `$^\wedge$' are described in our companion paper \cite{Nguyen12a}.

\begin{figure*}[ht]
\centering
\includegraphics[width=12cm]{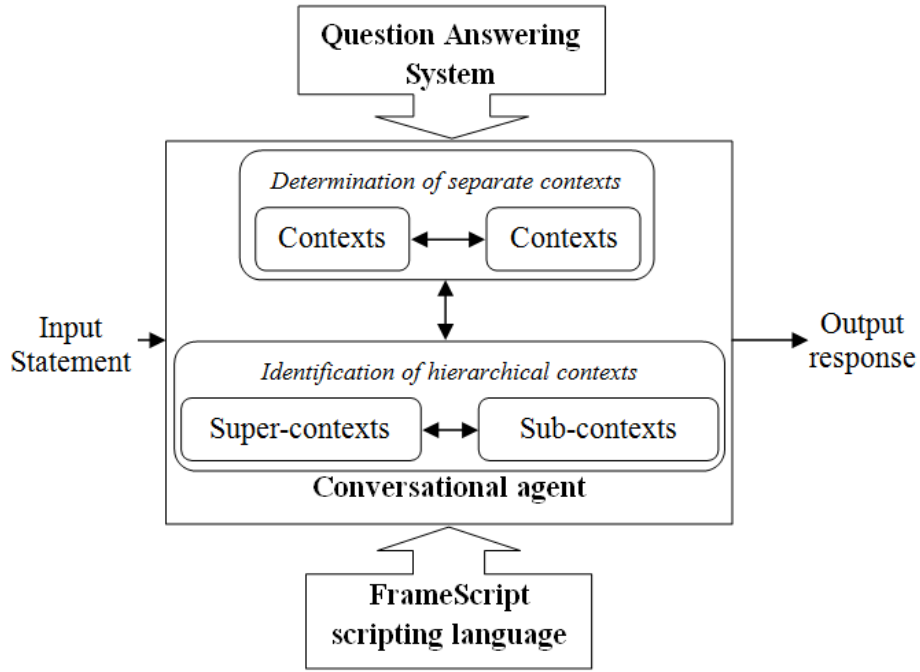}
\caption{Architecture of our Vietnamese text-based conversational agent.}
\label{Figure3}
\end{figure*}

\section{Text-based Conversational Agent for Vietnamese}\label{Ourapproach}

The architecture of our text-based conversational agent is shown in figure \ref{Figure3}, in which our Vietnamese question answering system is similar to VnQAS \cite{Nguyen09}.

Our question answering system consists of two components: the Natural language question analysis and the Answer retrieval. 
The communication between two components is an intermediate representation of the input question. 
If our Vietnamese question answering system is unable to give an answer for a user's question, the question will be considered as the input for our conversational agent. 
Our contribution focuses on presenting a Vietnamese text-based conversational agent as a backup component to provide necessary information related to the user's question.

The architecture enables the construction of hierarchical rules for conversational contexts using FrameScript language \cite{McGill:2003-8}. 
There are two steps to build these conversational contexts hierarchically: \textit{determining separate contexts} and \textit{identifying hierarchical contexts}. 
The first step identifies all possible contexts as well as the transitions between contexts.  
The next step is used to organize these contexts into a hierarchy which can handle unexpected inputs.

\subsection{Determining separate contexts}

\begin{table}[ht]
\caption{Script examples of ``subjects''}
\label{tab:1}
\centering
\begin{tabular}{l}
\hline
mon\_hoc\_tien\_quyet ::\\
trigger\{* môn học tiên quyết$_{prerequisite\ subject}$ *\} \\
* môn học tiên quyết$_{prerequisite\ subject}$ * ==$>$\\
$[$ \textit{\textbf{Môn học tiên quyết}} của một môn học là môn học bắt buộc sinh viên phải\\
hoàn thành trước khi học môn học đó. Bạn có muốn biết thêm thông tin về\\
\textit{\textbf{môn học điều kiện}}?\\
$_{The\ \textbf{prerequisite}\ \textbf{subject}\ of\ a\ subject\ is\ a\ imperative\ subject\ that\ students\ must\ have}$\\
$_{completed\ before\ learning\ that\ subject.\ Do\ you\ want\ to\ know\ about\ the\ \textbf{conditional}\ \textbf{subject}?}$ $]$\\
\{* Có$_{Yes}$ * | * có$_{yes}$ * \} ==$>$ \\
$[$ $\#$goto(mon\_hoc\_dieu\_kien, $<<$* môn học điều kiện$_{conditional\ subject}$ *$>>$) $]$\\
\{* Không$_{No}$ * | * không$_{no}$ * \} ==$>$ \\
$[$ Bạn có muốn biết thêm thông tin về khóa luận? $\#$goto(khoa\_luan)\\
$_{Do\ you\ want\ to\ know\ about\ the\ thesis?}$ $]$\\
;;\\
\\
mon\_hoc\_dieu\_kien ::\\
trigger\{* môn học điều kiện$_{conditional\ subject}$ *\}\\
* môn học điều kiện$_{conditional\ subject}$ * ==$>$\\
$[$ \textit{\textbf{Môn học điều kiện}} là các môn học giáo dục thể chất, giáo dục quốc phòng\\
-- an ninh và kỹ năng mềm. Bạn có muốn biết thêm thông tin về \textit{\textbf{khóa luận}}?\\
$_{The\ \textbf{conditional}\ \textbf{subjects}\ are\ subjects\ such\ as\ health\ education, national\ defence\ education}$\\
$_{,\ and\ soft\ skills. Do\ you\ want\ to\ know\ about\ the\ \textbf{thesis}?}$ $]$\\
\{* Có$_{Yes}$ * | * có$_{yes}$ * \} ==$>$ $[$ $\#$goto(khoa\_luan) $]$\\
\{* Không$_{No}$ * | * không$_{no}$ * \} ==$>$\\
$[$ Bạn muốn biết thông tin gì?$_{What\ information\ do\ you\ want\ to\ know\ ?}$ $]$\\
;;\\
\hline
\end{tabular}
\end{table}

Given an input statement, the system will identify the current context.
The current context is determined by matching pattern expressions of rules to produce associated responses, and to possibly transform to a next context or turn to a previously visited context.

The hierarchical rules are organized into contexts, therefore, the transformation among contexts are based on the change of states from a script to other scripts by using the transformation rules. 
These transformations depend on the first step aiming to control the conversational stream into the specific knowledge domain. In table \ref{tab:1}, scripts  \textit{``mon\_hoc\_tien\_quyet''} and \textit{``mon\_hoc\_dieu\_kien''} represent contexts of \textit{``môn học tiên quyết$_{prerequisite\ subject}$''} and \textit{``môn học điều kiện$_{conditional\ subject}$''} respectively. Both of example scripts representing the corresponding contexts have a \textit{trigger} to identify a matched input to become current context. 
These scripts contain the transformation rules in the form of \textit{$\#$goto(a$\_$script)} to change the state of current context.

When our text-based conversational agent encounters user's question:
\textit{``môn học tiên quyết là gì?''} (\textit{``What is the prerequisite subject?''}). 
The trigger of script \textit{``mon\_hoc\_tien\_quyet''} will process this input, and context \textit{``môn học tiên quyết $_{prerequisite\ subject}$''} will become the current context. 
If user replies with \textit{``Có$_{Yes}$''} for the question:
 
\textit{``Bạn có muốn biết thêm thông tin về môn học điều kiện?''} 

\textit{``Do\ you\ want\ to\ know\ about\ the\ conditional subject?''}

the response expression in transformation rule will perform changing the current context to the context \textit{``môn học điều kiện$_{conditional\ subject}$''}.
In table \ref{tab:2}, we have an example of the transformation between these contexts. 

\begin{table}[ht]
\caption{Transformations between contexts}
\label{tab:2}
\centering
\begin{tabular}{l}
\hline
\textbf{User:} \textit{\textbf{môn học tiên quyết}} là gì? $_{What\ is\ the\ \textbf{prerequisite}\ \textbf{subject}?}$\\
\textbf{CA:} \textit{\textbf{Môn học tiên quyết}} của một môn học là môn học bắt buộc sinh viên phải\\
hoàn thành trước khi học môn học đó. Bạn có muốn biết thêm thông tin về\\
\textit{\textbf{môn học điều kiện}}? 
\\
$_{The\ \textbf{prerequisite}\ \textbf{subject}\ of\ a\ subject\ is\ a\ imperative\ subject\ that\ students\ must\ have}$\\
$_{completed\ before\ learning\ that\ subject.\ Do\ you\ want\ to\ know\ about\ the\ \textbf{conditional}\ \textbf{subject}?}$\\
\\
\textbf{User:} \textit{\textbf{Có}}. Tôi muốn biết.$_{Yes.\ I\ want\ to\ know.}$\\
\textbf{CA:} \textit{\textbf{Môn học điều kiện}} là các môn học giáo dục thể chất, giáo dục quốc phòng\\
-- an ninh và kỹ năng mềm. Bạn có muốn biết thêm thông tin về \textit{\textbf{khóa luận}}?\\
$_{The\ \textbf{conditional}\ \textbf{subjects}\ are\ subjects\ such\ as\ health\ education,\ national\ defence\ education,}$\\
$_{\ and\ soft\ skills.\ Do\ you\ want\ to\ know\ about\ the\ \textbf{thesis}?}$\\
\\
\textbf{User:} Tôi không muốn biết về nó.$_{No,\ I\ do\ not.}$\\
\textbf{CA:} Bạn muốn biết thông tin gì?$_{What\ information\ do\ you\ want\ to\ know\ ?}$\\
\hline
\end{tabular}
\end{table}

\begin{table}[ht]
\caption{Order of transformation rules}
\label{tab:3}
\centering
\begin{tabular}{l}
\hline
mon\_hoc ::\\
trigger\{* môn học$_{subject}$ *\}\\
* môn học tiên quyết$_{prerequisite\ subject}$ * ==$>$ \\
\ \ \ \ \ $[$ $\#$goto(mon\_hoc\_tien\_quyet, $^\wedge$0) $]$\\
* môn học điều kiện$_{conditional\ subject}$ * ==$>$ \\
\ \ \ \ \ $[$ $\#$goto(mon\_hoc\_dieu\_kien, $^\wedge$0) $]$\\
* môn học$_{subject}$ * ==$>$ \\
$[$ Các loại \textbf{môn học} gồm có các môn học bắt buộc, các môn học tự chọn,\\
các môn học tiên quyết của một môn học, các môn học điều kiện và khóa luận.\\
$_{The\ \textbf{subjects}\ consist\ of\ imperative\ subjects,\ optional\ subjects,\ prerequisite\ subjects,}$\\
$_{conditional\ subjects\ and\ thesis.}$ $]$\\
;;\\
\\
quy\_che\_dao\_tao ::\\ 
\ \ \ \ \ * môn học tiên quyết$_{prerequisite\ subject}$ * ==$>$  \ \ \ \ \ $//$Rule 1\\
\ \ \ \ \ \ \ \ \ \ $[$ $\#$goto(mon\_hoc\_tien\_quyet, $^\wedge$0) $]$\\
\ \ \ \ \ * môn học điều kiện$_{conditional\ subject}$ * ==$>$  \ \ \ \ \ $//$Rule 2\\
\ \ \ \ \ \ \ \ \ \ $[$ $\#$goto(mon\_hoc\_dieu\_kien, $^\wedge$0) $]$\\
\ \ \ \ \ * môn học$_{subject}$ * ==$>$ $[$ $\#$goto(mon\_hoc, $^\wedge$0) $]$  \ \ \ \ \ $//$Rule 3\\ 
;;\\
\hline 
\end{tabular}
\end{table}

\subsection{Identifying hierarchical contexts}

After determining the separate contexts, contexts would be arranged into a hierarchy to handle unexpected inputs.
Therefore, the second step is used to identify the relation among contexts. Specifically, this is the relationship between super-context and its sub-contexts.
A context is the super-context of other contexts if the transformation rule for this context is placed after the transformation rules for the sub-contexts. 
This aims to recognize the suitable contexts to satisfy input statements.  

In table \ref{tab:3}, we have the context \textit{``môn học\ $_{subject}$''} represented by script \textit{``mon\_hoc''}, and its two sub-contexts \textit{``môn học tiên quyết$_{prerequisite\ subject}$''} and \textit{``môn học điều kiện $_{conditional\ subject}$''}. 
Assuming context \textit{``quy chế đào tạo $_{academic\ regulation}$''} described by script \textit{``quy\_che\_dao\_tao''} is being the current context,
and the Rule 3 is placed in front of both Rule 1 and Rule 2 in script \textit{``quy\_che\_dao\_tao''}, with the input statement:  
\textit{``môn học tiên quyết là gì $_{What\ is\ the\ prerequisite\ subject}$ ?''} or 
\textit{``môn học điều kiện là gì $_{What\ is\ the\ conditional}$ $_{subject}$ ?''}, 
the current context is always moved to context \textit{``môn học$_{subject}$''} without transforming to appropriate contexts. 
Therefore, the output response providing definition of \textit{``môn học$_{subject}$''} does not satisfy user's requirement.

Consequently, the transformation rule for super-context must be constructed to follow transformation rules for its sub-contexts as shown in the script \\ \textit{``quy\_che\_dao\_tao''} in table \ref{tab:3}. The transcript of dialogue described in table \ref{tab:4} illustrates the ordered transformation among those contexts.

\begin{table}[ht]
\caption{Ordered transformation between contexts}
\label{tab:4}
\centering
\begin{tabular}{l}
\hline
\textbf{User:} Các loại \textit{\textbf{môn học}} trong giảng dạy tín chỉ?\\
$_{Which\ kinds \ of\ \textbf{subjects}\ are\ in\ the\ credit-based\ teaching?}$\\
\textbf{CA:} Các loại \textit{\textbf{môn học}} gồm có các môn học bắt buộc, các môn học tự chọn,\\
các môn học tiên quyết của một môn học, các môn học điều kiện và khóa luận.\\
$_{The\ kinds\ of \textbf{subjects}\ consist\ of\ imperative\ subjects,\ optional\ subjects,\ prerequisite\ subjects,}$\\
$_{conditional\ subjects\ and\ thesis.}$\\
\\
\textbf{User:} \textit{\textbf{môn học điều kiện}} là gì?$_{What\ is\ the\ \textbf{conditional}\ \textbf{subject}?}$\\
\textbf{CA:} \textit{\textbf{Môn học điều kiện}} là các môn học giáo dục thể chất, giáo dục quốc phòng\\
-- an ninh và kỹ năng mềm. Bạn có muốn biết thêm thông tin về \textit{\textbf{khóa luận}}?\\
$_{The\ \textbf{conditional}\ \textbf{subjects}\ are\ subjects\ such\ as\ health\ education,\ national\ defence\ education,}$\\
$_{\ and\ soft\ skills.\ Do\ you\ want\ to\ know\ about\ the\ \textbf{thesis}?}$\\
\\
\textbf{User:} Tôi không muốn biết về nó.$_{No,\ I\ do\ not.}$\\
\textbf{CA:} Bạn muốn biết thông tin gì?$_{What\ information\ do\ you\ want\ to\ know\ ?}$\\
\hline
\end{tabular}
\end{table}

\section{Experiments and Discussion}\label{Experiment}
\subsection{Experimental results\\for Vietnamese text-based conversational agent}

For this experiment, we built conversational interactions of 16 contexts from a chapter in the academic regulations of the Vietnam National University, Hanoi. 
Our goal is to support students understanding the academic regulations of the university via a question answering system.

\begin{table}[ht]
\caption{List of transformations among contexts}
\label{tab:transfer}
\centering
\begin{tabular}{|p{3cm}@{\quad} | p{8.5cm}|}
\hline \textbf{Context} & \textbf{Transferred contexts} \\
\hline quy định$_{regulation}$ & hình thức dạy học$_{teaching\ form}$, tín chỉ$_{credit}$ \\
\hline		hình thức dạy học $_{teaching\ form}$ & lên lớp$_{on\ class}$, thực hành$_{practice}$,\\
		& tự học bắt buộc$_{imperative\ self-study}$\\
\hline		lên lớp$_{on\ class}$ & thực hành$_{practice}$, tự học bắt buộc $_{imperative\ self-study}$\\
\hline		thực hành$_{practice}$ & tự học bắt buộc$_{imperative\ self-study}$, lên lớp$_{on\ class}$\\
\hline		tự học bắt buộc $_{imperative\ self-study}$ & lên lớp$_{on\ class}$, thực hành$_{practice}$\\
\hline		tín chỉ$_{credit}$ & chương trình đào tạo $_{training\ program}$, giờ tín chỉ$_{credit\ hour}$\\
\hline		chương trình đào tạo $_{training\ program}$ & hình thức đào tạo$_{training\ form}$ \\
\hline		giờ tín chỉ$_{credit\ hour}$ & tín chỉ$_{credit}$, môn học$_{subject}$ \\
\hline		môn học$_{subject}$ & môn học bắt buộc$_{imperative\ subject}$, \\
		& môn học tự chọn $_{optional\ subject}$,\\
		& môn học tiên quyết$_{prerequisite\ subject}$,\\
		& môn học điều kiện$_{conditional\ subject}$, \\
		& chương trình đào tạo$_{training\ program}$, \\
		& khóa luận$_{thesis}$\\
\hline
\end{tabular}
\end{table}

Table \ref{tab:transfer} describes the transformations among contexts.
The conversational interactions between users and our Vietnamese text-based conversational agent start at the default context considered as current context named \textit{``quy chế đào tạo$_{academic\ regulation}$''}. 
The following context \textit{``hình thức dạy học$_{teaching\ form}$''} gives students the common information about teaching forms in the university.
Then contexts of \textit{``tự học bắt buộc$_{imperative\ self-study}$''}, \textit{``lên lớp$_{on\ class}$''} and \textit{``thực hành$_{practice}$''} aim to detail kinds of teaching forms.
Table \ref{tab:transfer} also shows that the conversational interactions may generate repeated transformations among contexts such as \textit{``giờ tín chỉ$_{credit\ hour}$''} and \textit{``tín chỉ$_{credit}$''}. Thus we use a method for logging these interactions in order to propose the change to another context.

In our experiment, we collected the inputs from 30 students interacting with our Vietnamese text-based conversational agent.
A session contains all communications between one student and the system. 
On average, a student has 14 interactions (inputs) with the system in a session to retrieve the desire information. 
We had in total 417 interactions from 30 students. 
When a response of user input correctly provides the desire information, this response is regarded as satisfying the requirement of students.
We achieved an accuracy of 79.4\% with 331 input statements having satisfying  response from our system.  

\begin{table}[ht]
\caption{Unsatisfying analysis}
\label{tab:missResults}
\centering
\begin{tabular}{l@{\quad} l}
\hline Reason &  Number of user inputs\\
\hline  Constructing of patterns is not appropriate & 75\\ 
		Organizing of hierarchical contexts is not compatible & 11\\
\hline
\end{tabular}
\end{table}

Table \ref{tab:missResults} presents the error analysis for the 86 inputs which their responses did not satisfy the students.
The causes came from the construction of patterns and the organization of hierarchical contexts.
It clearly shows that most cases come from constructing patterns of rules.
This could be easily rectified by refining or adding more script rules.
Table \ref{tab:degree} shows the students' degree of satisfaction when interacting with our conversational agent.
We provided a scale of 1 to 5 for 30 students to separately evaluate based on the information provided by our text-based conversational agent, that is, 1: bad, 2: normal, 3: good, 4: very good, 5: excellent.
The feedback is that most students find the system interesting and highly value the system because of its practical use. 
To the best of our knowledge, this is the first text-based conversational agent for Vietnamese.
\begin{table}[ht]
\caption{The satisfied degree of students}
\label{tab:degree}
\centering
\begin{tabular}{l@{\quad} l@{\quad} l@{\quad} l@{\quad} l@{\quad} l@{\quad}}
\hline Degree of satisfaction & 1 & 2 & 3 & 4 & 5\\
\hline Number of students & 3 & 1 & 13 & 9 & 4\\
\hline
\end{tabular}
\end{table}

\subsection{Discussion} 
\label{Discussion}
Because constructing rules depends on the identification of super-contexts and their sub-contexts in Vietnamese, so it causes difficulties in designing the hierarchy of contexts. 
Consequently, we want to simplify this designing phase according to the process of semantic knowledge acquisition. 
We built additional scripts as shown in our companion paper \cite{Nguyen12a} to detect  noun phrases, question phrases and relation phrases or semantic constraints between them for Vietnamese. 
Using these scripts, we constructed pattern expressions and got the suitable phrases from response expressions. 
These phrases actually are keywords which may be used as patterns of rules in the hierarchical contexts.

In addition, our Vietnamese text-based conversational agent is integrated with our ontology-based Vietnamese question answering system \cite{Nguyen09} to form a general system. 
Our goal is to retrieve the necessary information from user's utterance to support our Vietnamese question answering system in providing answers to users.
We consider the process that the Answer retrieval component similarly measures between elements of the intermediate representation of user's question and the ontology's elements. 
In case of ambiguity for the similarity among ontology's elements is still present, the system will interacts with the users by presenting different options to get the correct ontology's elements.
In this process, we can construct the supplemental scripts to solve ambiguities from ontology knowledge base. 
Using these scripts, we can retrieve the suitable elements from ontology through the conversational contexts structured based on the given ontology.

\section{Conclusion}\label{Conclusion}

In this paper, we proposed a Vietnamese text-based conversational agent architecture as backup component integrated with our Vietnamese question answering system to form a general system.
We focused on presenting an approach to construct the hierarchical contexts consisting of organized rules over a specific knowledge domain.
There are two steps to construct the conversation contexts: the first step to identify the transformations from a context to other contexts, and the second step to organize these contexts into a hierarchy to handle unexpected inputs.
Our contribution is to provide the suitable information related to users' statements and to retrieve the necessary knowledge to support our question answering system in providing answers.

The experimental results are promising, with positive evaluation from users for our Vietnamese text-based conversational agent. 
To the best of our knowledge, this is the first Vietnamese text-based conversational agent to enable users to interact with the system via a natural language interface.

In the future, our text-based conversational agent will be extended not only to communicate with users but also to get the necessary information related to ontology knowledge base from input utterances. 
We will build scripts to resolve the ambiguity between elements of ontology such as the similarity of string names among classes or instances in the ontology. 
The constructed scripts would be utilized to generate options in order to obtain terms from conversational interactions with users.

\section*{Acknowledgements}

This work is partially supported by the Research Grant from Vietnam National University, Hanoi No. QG.10.39.

The authors would like to acknowledge Vietnam National Foundation for Science and Technology Development (NAFOSTED) for their financial support to present the work at the conference.

\bibliographystyle{splncs03}
\bibliography{references}

\end{document}